# COMBINING NEURAL NETWORKS FOR SKIN DETECTION

Chelsia Amy Doukim[1], Jamal Ahmad Dargham[1], Ali Chekima[1] and Sigeru Omatu[2]

[1]School of Engineering and Information Technology, Universiti Malaysia Sabah, Malaysia
`chelsia_amy@hotmail.com,jamalad@ums.edu.my,chekima@ums.edu.my`
[2]Computer and Systems Sciences, Graduate School of Engineering, Osaka Prefecture University, Sakai, Osaka 599-8531, Japan
`omatu@cs.osakafu-u.ac.jp`

## ABSTRACT

*Two types of combining strategies were evaluated namely combining skin features and combining skin classifiers. Several combining rules were applied where the outputs of the skin classifiers are combined using binary operators such as the AND and the OR operators, "Voting", "Sum of Weights" and a new neural network. Three chrominance components from the $YC_bC_r$ colour space that gave the highest correct detection on their single feature MLP were selected as the combining parameters. A major issue in designing a MLP neural network is to determine the optimal number of hidden units given a set of training patterns. Therefore, a "coarse to fine search" method to find the number of neurons in the hidden layer is proposed. The strategy of combining $C_b/C_r$ and $C_r$ features improved the correct detection by 3.01% compared to the best single feature MLP given by $C_b$-$C_r$. The strategy of combining the outputs of three skin classifiers using the "Sum of Weights" rule further improved the correct detection by 4.38% compared to the best single feature MLP.*

## 1. INTRODUCTION

Skin detection is an important preliminary process for subsequent feature extraction in a wide range of image processing techniques such as face detection, face tracking, gesture analysis, content-based image retrieval systems, various computer vision applications, etc. Studies have shown that by combining more than one feature or classifier, the performance of the skin detection system is improved. Zhu *et al.* [1] combined two Gaussian feature spaces where the first one is related to the colour distribution and the second one is related to the skin spatial and shape distribution. The combined feature method performed better compared to single feature and also other generic skin model namely histogram model, single Gaussian model and Gaussian Mixture model. Brand and Mason [2] evaluated the performance of the combined colour components or features from the RGB colour space and concluded that the combination of (R/G + R/B + G/B) gave better performance than the single colour component. Jiang *et al.* [3] proposed a Skin Probability Map (SPM) based skin detection system that integrated the colour, texture and space information and claimed that their proposed method performed better than the generic SPM method. Gasparini *et al.* [4] combined different skin classifiers based on different colour features using several combination rules such as the sum rule, the product rule, the majority rule and the author's proposed skin corrected by non-skin (SCNS) rule. The performance was evaluated in terms of recall and precision. The performance for all combining rules increases in terms of precision compared to single classifier and the most precision-oriented is given by the product rule. Sajedi *et al.* [5] combined a block-based skin detection classifier with a boosted pixel-based classifier. The boosted pixel-based classifier is modified by combining several explicit boundary skin classifiers based on different colour features. The authors claimed that their method is more robust to variations of skin colour compared to Self-organizing Map (SOM), Fuzzy Integral, conventional pixel-based method and Bayesian network approach.

Neural networks have been used successfully as skin classifier. However, there is no research has been done for combining the neural network-based skin classifiers. In this paper, the multi-layer perceptron (MLP) neural network is used for skin detection. Several chrominance components from the $YC_bC_r$ colour space are used as the skin colour features. Several strategies for combining MLP neural networks for skin detection are proposed and their performance on skin detection is evaluated. The paper is organised as follows. Section 2 briefly describes the data preparation. In Section 3, the neural network properties used in this work are explained. Section 4 explains the method for finding the number of neurons in the hidden layer. Section 5 describes the performance metrics used to evaluate the skin detection performance and the performance for single feature MLP is given. Section 6 explains the combining skin features strategy. In Section 7, the strategy of combining skin classifiers using several combining rules are described. Finally, Section 8 concludes the paper.

## **2. DATA PREPARATION**

The database used in this work is the Compaq database [6]. This database consists of 13,640 images with its corresponding masked images. These images contain skin pixels belonging to persons of different origins, with unconstrained illumination and background conditions, which make the skin detection task more challenging and difficult. Figure 1 shows an example of images and their corresponding masked images. Two sets of data are prepared namely the training data and test data. The training data comprises training and validation samples that will be used to train the MLP neural networks. The training sample consists of 420,000 image pixels and will be validated by a similar number of image pixels randomly selected from the Compaq database. Note that each image pixel can be selected only once.  The training data is divided into 30 data files where each data file consists of 14,000 pixels from the training sample and an equal number of pixels from the validation sample. Each data file will be used to train a network during a training run. The test data consists of 100 images selected at random from the Compaq database. The test images are used to evaluate the performance of the skin detection system.

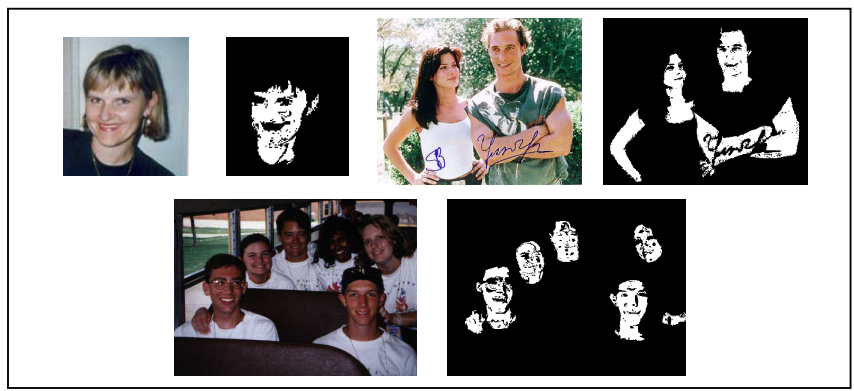

Figure 1.  Example of images from the Compaq database with their corresponding masked images.

## 3. NEURAL NETWORK PROPERTIES

One of the important aspects in designing a MLP neural network is how to determine the network topology. The input size is dictated by the number of features of available inputs and the output size is dictated by the number of classes. Thus, the two decisions that must be made regarding the hidden units are to determine the number of hidden layers and the number of neurons in each hidden layer. Fu [7] stated that using only one hidden layer is sufficient to solve many practical problems and thus, one hidden layer MLP neural network is used in this work. The determination of the number of neurons in the hidden layer will be discussed in Section 4. Hence, the neural network topology will be C-HN-O, where C indicates the input neuron which is the chrominance component, HN is the number of neurons in the hidden layer and O is the output neuron. The output layer will have one neuron decoded as 1 for skin and 0 for non-skin. The training algorithm used is the Levenberg-Marquardt because of its rapid convergence time compared with other fast training algorithms such as Conjugate gradient and Quasi-Newton. The transfer function applied is the sigmoid function. The maximum number of epochs is set to 500 as it is proven through trials and error that the number of epochs required for convergence using the Levenberg-Marquardt training algorithm always occurs well below the 500 epochs [8]. The training goal selected is $1 \times 10^{-6}$

## 4. DETERMINATION OF THE NUMBER OF NEURONS IN THE HIDDEN LAYER

The first step is to find the number of neurons for several chrominance components in the $YC_bC_r$ colour space, namely, $C_b$, $C_r$, $C_b/C_r$, $C_b.C_r$ and $C_b-C_r$. Each chrominance is treated as an input neuron of a MLP neural network. Thus, the MLP neural network structure for each chrominance component is 1-HN-1. Existing technique for finding the number of neurons in the hidden layer are network growing [9] and network pruning [10]. In this work, a modified network growing called "coarse to fine search" method is applied. This method iss divided into two stages. The first stage is a coarse search using binary search. Hence, the values of HN to be tested are 1, 2, 4, 8, 16, 32, 64 and 128. The maximum number of HN is set at 128 because a larger number of HN requires more memory space. Each network structure with a given HN is trained 30 times (training runs) using different initial values and training and validation data and its average Mean Squared Errors (MSE) over the 30 runs is calculated. The HN value that gives the lowest MSE will be selected. In the second stage, a sequential search or fine search around a fixed range of the chosen HN from the first stage is done in order to obtain the optimal number of neurons in the hidden layer. Assuming that the chosen HN from the first stage is $HN_B$, then the nearest HN values by that $HN_B$ are $HN_{BL}$ for the nearest lower value of $HN_B$ and $HN_{BH}$ for the nearest upper value of $HN_B$. The range of the sequential search is fixed from $HN_{BL} + [(1/2) \times (HN_B-HN_{BL})]$ to $HN_B + [(1/2) \times (HN_{BH}-HN_B)]$. However, this condition is true only if the $HN_B$ equals to 4, 8, 16, 32 or 64. If $HN_B$ equals to 2, then the sequential search will begin from $HN_{BL}$ to $HN_B + [(1/2) \times (HN_{BH}-HN_B)]$. For the case when $HN_B$ equals to 128, the range of the sequential search begins from $HN_{BL} + [(1/2) \times (HN_B–HN_{BL})]$ to $HN_B$, since 128 is the fixed maximum HN value. As can be seen from Figure 2, the lowest HN that gives the lowest MSE=0.0426 which is HN=64 is chosen. The $HN_{BL}$ and $HN_{BH}$ are 32 and 128 respectively. Thus, the sequential search range will be from HN=48 to HN=96. The result of the sequential search for $C_b$ chrominance component is shown in Figure 3. As can be seen from Figure 3, the MSE results are consistent for few HN values at first and then the variation occurs until it gives the lowest MSE value of 0.0425. The lowest HN that gives the lowest MSE is HN = 91, thus, the fixed network structure for $C_b$ chrominance is one input neuron, 91 neurons in the hidden layer and one output neuron, or 1-91-1. The proposed algorithm is found to be effective in terms of training time since the binary search method was applied rather than using the sequential search method alone. Table 1 gives the fixed network structure for every chrominance component.

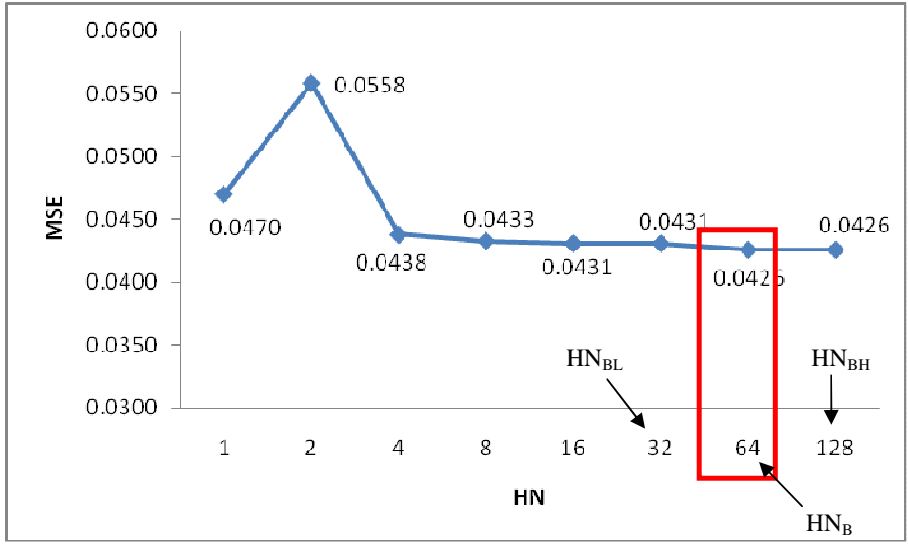

Figure 2. Coarse (binary) search for $C_b$ chrominance component.

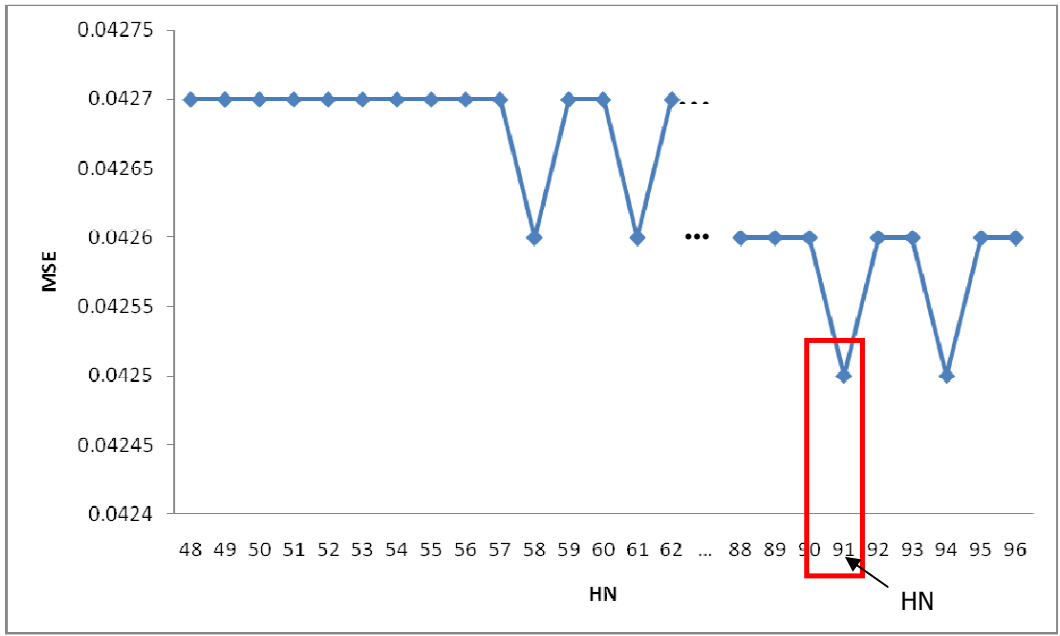

Figure 3. Fine (Sequential) search for $C_b$ chrominance component.

Table 1. Fixed network structure for every chrominance component.

| Network Structure | Chrominance Component | Network Structure (C-HN-O) |
|---|---|---|
| Single Input | $C_b$ | 1-91-1 |
| | $C_r$ | 1-96-1 |
| | $C_b/C_r$ | 1-128-1 |
| | $C_b.C_r$ | 1-96-1 |
| | $C_b-C_r$ | 1-112-1 |

## 5. SKIN DETECTION

The fixed MLP neural networks are used to segment all the images in the test dataset into skin and non-skin regions. To evaluate the performance of each neural network, three performance metrics are used. The first metric is the correct detection rate, **CDR** and is given in Equation 1. The false acceptance rate **FAR** is the percentage of identification instances in which false acceptance occurs. For example, an unauthorized person is identified as an authorized one. The false rejection rate **FRR** is the percentage of identification instances in which false rejection occurs. This is the case when the system fails to recognize an authorized person and rejects that person as an impostor. The FAR and FRR are expressed in Equations (2) and (3), respectively.

$$CDR = \frac{\text{Number of pixels correctly classified}}{\text{Total pixels in the test dataset}} \quad (1)$$

$$FAR = \frac{\text{Number of non-skin pixels classified as skin pixels}}{\text{Total pixels in the test dataset}} \quad (2)$$

$$FRR = \frac{\text{Number of skin pixels classified as non-skin pixels}}{\text{Total pixels in the test dataset}} \quad (3)$$

Since the transfer function used is a sigmoid function, the MLP neural network will be producing the output between 0 and 1. Thus, the output of the neural network needs to be modified so that it is either 0 or 1. In this work, a single threshold value is used in determining the skin and non-skin classes. If the MLP network output is higher than the threshold value, then the output is 1. Otherwise, the output is 0. The threshold value used in this work is 0.5. As can be seen from Table 2, $C_b$-$C_r$ gives the highest correct detection rate.

Table 2. Skin detection performance for every MLP network structures.

| Network Structure | Chrominance Component | Correct Detection Rate (CDR) | False Acceptance Rate (FAR) | False Rejection Rate (FRR) |
|---|---|---|---|---|
| Single Input | $C_b$ | 70.81 | 28.34 | 0.85 |
| | $C_r$ | 76.18 | 23.48 | 0.34 |
| | $C_b/C_r$ | 78.63 | 20.50 | 0.86 |
| | $C_b.C_r$ | 59.66 | 35.02 | 5.31 |
| | **$C_b$-$C_r$** | **79.60** | **19.55** | **0.85** |

## 6. COMBINING SKIN FEATURES

For this strategy, two and three chrominance components or skin features are combined in the input layer. Hence, the three chrominance components that give the highest correct detection on their single feature MLP are selected namely $C_b$-$C_r$, $C_b/C_r$ and $C_r$. The number of neurons in the hidden layer or HN is determined using "coarse to fine search" technique as discussed earlier. The training parameters used such as training algorithms, transfer function, number of epochs and training goal are the same as defined in Section 3. Table 3 gives the fixed MLP neural network structures for every possible combination. The fixed MLP structures are used to segment the test image. The performance of the skin detection for every combining feature is shown in Table 4. The highest correct detection is given by combination of $C_b/C_r$ and $C_r$ features, with an improvement of 3.01% compared to the best single feature MLP given by $C_b$-$C_r$.

Table 3. Fixed network structure for every possible combination of features.

| Network Structure | Combination of Features | Network Structure (C-HN-O) |
|---|---|---|
| Two Inputs | $C_b$-$C_r$ & $C_b/C_r$ | 2-17-1 |
| | $C_b$-$C_r$ & $C_r$ | 2-114-1 |
| | $C_b/C_r$ & $C_r$ | 2-123-1 |
| Three Inputs | $C_b$-$C_r$, $C_b/C_r$ & $C_r$ | 3-5-1 |

Table 4. Skin detection performance for every combining feature.

| Network Structure | Combining Feature | Correct Detection Rate (CDR) | False Acceptance Rate (FAR) | False Rejection Rate (FRR) |
|---|---|---|---|---|
| Two Inputs | $C_b$-$C_r$ & $C_b/C_r$ | 79.03 | 20.15 | 0.82 |
| | $C_b$-$C_r$ & $C_r$ | 63.86 | 33.97 | 2.16 |
| | **$C_b/C_r$ & $C_r$** | **82.61** | **14.66** | **2.73** |
| Three Inputs | $C_b$-$C_r$, $C_b/C_r$ & $C_r$ | 73.61 | 25.67 | 0.72 |

## 7. COMBINING SKIN CLASSIFIERS

Four combining rules are used to combine the skin classifiers namely the binary operators AND and OR, Voting rule, Sum of Weights rule and a new neural network. The three chrominance components that gave the highest correct detection on their respective single feature MLP are selected for these combination strategies. Thus, the three chrominance components are $C_b$-$C_r$, $C_b/C_r$ and $C_r$.

### 7.1. Combination of Outputs of Skin Classifiers Using the AND and OR Operators

The outputs of two and three skin classifiers are combined using the AND and OR operators. For combining two skin classifiers, the possible combinations are $C_b$-$C_r$ and $C_b/C_r$, $C_b$-$C_r$ and $C_r$, and $C_b/C_r$ and $C_r$. Figure 4 illustrates the procedure of skin detection for combining the outputs of two skin classifiers using the binary operators. The skin classifiers, for example the $C_b$-$C_r$ and $C_b/C_r$, are used to segment 100 test images into skin and non-skin regions. The outputs from the respective skin classifier are thresholded and combined using the AND operator. A single threshold value of 0.5 is used for the skin and non-skin classification.

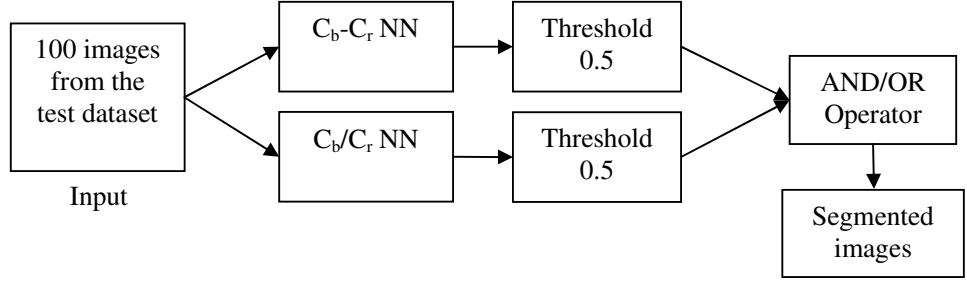

Figure 4. Procedure for combining the outputs of two skin classifiers using the binary operators.

### 7.2. Combination of Outputs of Skin Classifiers Using the Voting Rule

The "Voting" rule used is based on majority rule, which means that when any two or all three skin classifiers produced the outputs of 1, then it is a skin. Likewise, when any two or all three skin classifiers produced the outputs of 0, then it is a non-skin. Figure 5 illustrates the procedure of skin detection for this strategy.

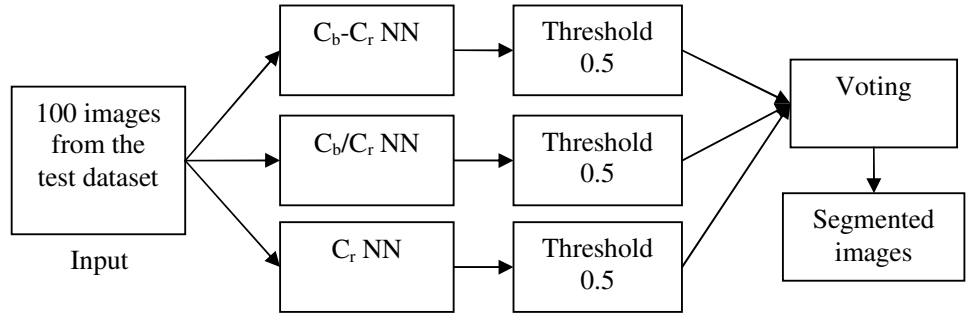

Figure 5. Procedure for combining the outputs of three skin classifiers using Voting rule.

### 7.3. Combination of Outputs of Skin Classifiers Using the Sum of Weights Rule

The weights for the three skin classifiers: $C_b$-$C_r$, $C_b/C_r$ and $C_r$, are fixed based on their correct detection rates. From Table 2, the correct detection rate for $C_b$-$C_r$, $C_b/C_r$ and $C_r$ are 79.60%, 78.63% and 76.18%, respectively. Thus, the weights for each neural network are fixed as expressed in Equation 4 to Equation 6.

$$W_{Cb\text{-}Cr} = 79.60/ (79.6+78.63+76.18) = 0.3396 \qquad (4)$$

$$W_{Cb/Cr} = 78.63/ (79.6+78.63+76.18) = 0.3354 \qquad (5)$$

$$W_{Cr} = 76.18/ (79.6+78.63+76.18) = 0.3250 \qquad (6)$$

where $W_{Cb\text{-}Cr} + W_{Cb/Cr} + W_{Cr} = 1$.

Figure 6 illustrates the procedure of skin detection using this strategy. The inputs that are fed into each neural network will be multiplied by their corresponding weights and produce the outputs $Y_1(i,j)$, $Y_2(i,j)$ and $Y_3(i,j)$, respectively. These outputs are then summed and thresholded in order to classify the skin and non-skin regions. Note that the threshold value used in this strategy is different because each neural network was multiplied by their corresponding fixed weights and thus each neural network will have different threshold values. Thus, the threshold value is determined empirically.

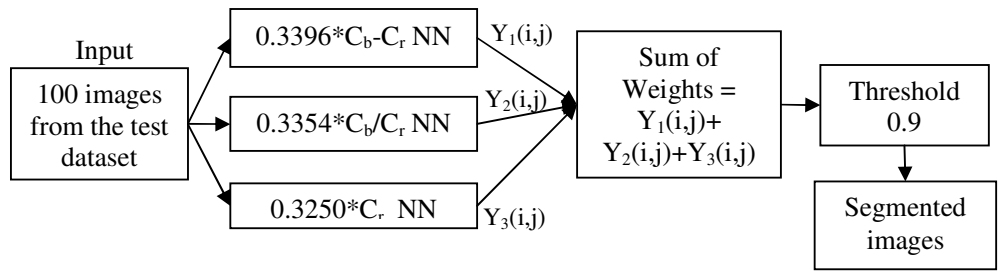

Figure 6. Procedure of combining the outputs of three skin classifiers using the Sum of Weights rule.

### 7.4. Combination of Outputs of Skin Classifiers Using a New Neural Network

A new 3-HN-1 MLP neural network is designed first before the combination strategy is carried out. The number of HN is determined using the "coarse to fine search" technique. The training data used is different from the previous designed MLP neural networks. Figure 7 illustrates the procedure for creating the training and validation samples for training the new neural network. For each neural network, a total of 420,000 pixels are selected at random to produce the training sample and a similar number of pixels are also selected at random to create the validation sample. The training parameters used are similar as those defined in Section 3. Thus, the fixed network structure is 3-126-1. Figure 8 illustrates the procedure for the combination strategy.

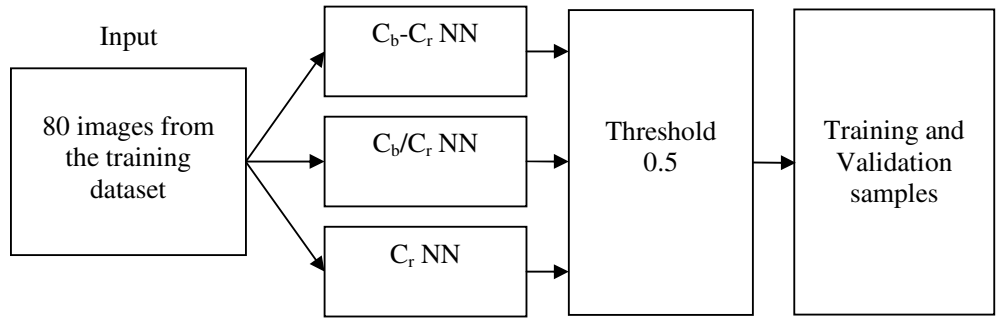

Figure 7. Procedure for creating the training and validation samples for training the new neural network.

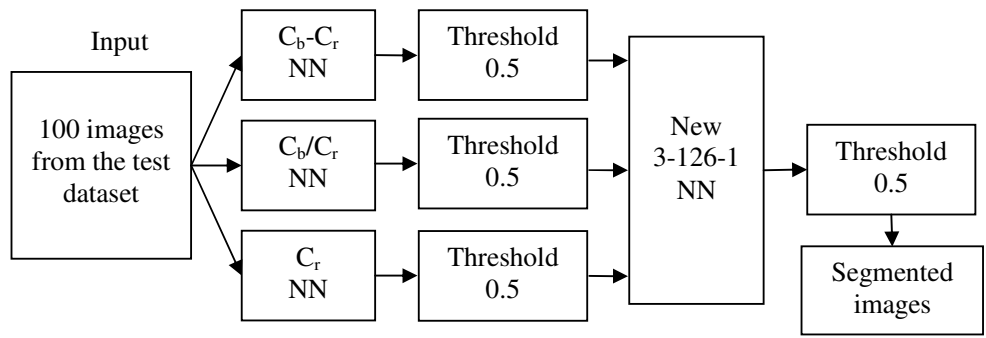

Figure 8. Procedure for combining the outputs of three skin classifiers using a new neural network.

### 7.5. Skin Detection Results and Analysis

Table 5 provides the skin detection performance in for every combining strategies applied to the skin classifiers. The best performance in terms of correct detection is given by the combination of three classifiers using the "Sum of Weights" rule. This strategy gives an improvement of 4.38% in terms of correct detection compared to the best single feature MLP given by $C_b$-$C_r$ feature. Furthermore, this strategy gives 1.37% more correct detection than the best combining skin features strategy given by the combination of $C_b/C_r$ and $C_r$ features. The results are compared with Bayes' rule classifier by Jones and Rehg [6] based on the skin detection performance using the Compaq database. Both best combining feature and combining classifier strategies give 2.61% and 3.98% more correct detection respectively compared to the Bayes' rule classifier by Jones and Rehg [6]. Figure 8 and Figure 9 show the best and the worst test images segmented using the best single feature MLP, the best combining feature MLP and the best combining classifier MLP respectively.

Table 5. Skin detection performance for several different combination strategies.

| Number of Outputs | Combination Rule/ Operator | Chrominance Component | CDR | FAR | FRR |
|---|---|---|---|---|---|
| Two Outputs | AND | $C_b$-$C_r$ & $C_b/C_r$ | 79.77 | 19.35 | 0.89 |
| | | $C_b$-$C_r$ & $C_r$ | 82.21 | 16.90 | 0.89 |
| | | $C_b/C_r$ & $C_r$ | 82.29 | 16.80 | 0.91 |
| | OR | $C_b$-$C_r$ & $C_b/C_r$ | 78.46 | 20.17 | 0.83 |
| | | $C_b$-$C_r$ & $C_r$ | 73.56 | 26.13 | 0.31 |
| | | $C_b/C_r$ & $C_r$ | 72.53 | 27.18 | 0.30 |
| Three Outputs | AND | $C_b$-$C_r$, $C_b/C_r$ & $C_r$ | 82.38 | 16.69 | 0.92 |
| | OR | | 72.53 | 27.18 | 0.30 |
| | Voting | | 79.50 | 19.66 | 0.84 |
| | **Sum of Weights** | | **83.98** | **14.90** | **1.12** |
| | New 3-126-1 Neural Network | | 82.21 | 16.90 | 0.89 |

## 8. CONCLUSION

In this work, several combination strategies for combining MLP neural networks for skin detection were evaluated. A modified network growing technique for finding the number of neurons in the hidden layer of a MLP neural network was applied. Three chrominance components $C_b$-$C_r$, $C_b/C_r$ and $C_r$ that gave the highest CDR on their respective MLP were used for the combination. The combination of $C_b/C_r$ and $C_r$ features improved the CDR by 3.01% compared to the best single feature MLP given by $C_b$-$C_r$. Combining classifier using Sum of Weights strategy further improved the CDR by 4.38% compared to the best single feature MLP. Furthermore, combining classifiers using the Sum of Weights strategy improved the correct detection rate by 3.98% compared to the Bayes' rule classifier reported by Jones and Rehg [6] using the Compaq database.

| Techniques | Original Image | Masked Image | Segmented Image |
|---|---|---|---|
| Single feature MLP | 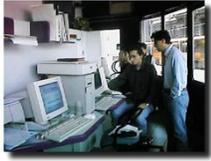 | 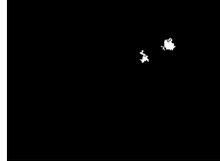 | 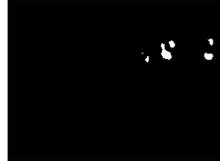 |
| Combining feature MLP | 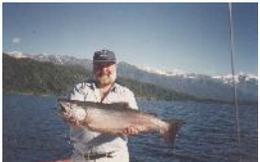 | 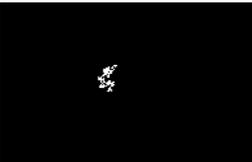 | 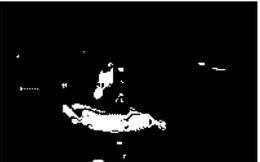 |
| Combining classifier MLP | 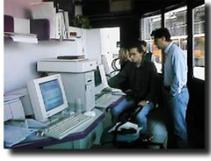 | 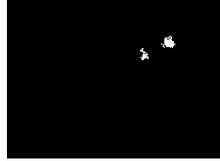 | 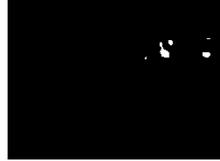 |

Figure 8. The best test images segmented using the best single feature MLP, the best combining feature MLP and the best combining classifier MLP.

| Techniques | Original Image | Masked Image | Segmented Image |
|---|---|---|---|
| Single feature MLP | 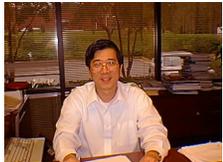 | 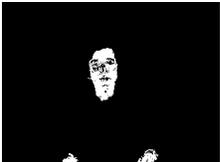 | 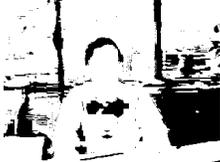 |
| Combining feature MLP | 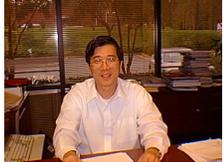 | 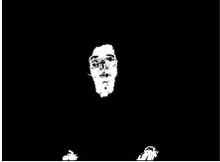 | 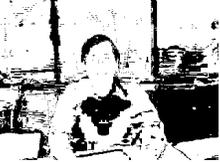 |
| Combining classifier MLP | 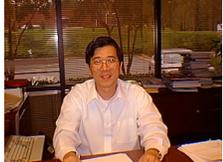 | 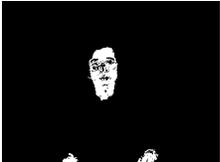 | 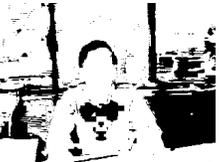 |

Figure 9. The worst test images segmented using the best single feature MLP, the best combining feature MLP and the best combining classifier MLP.